\documentclass[twoside,11pt]{article}

\usepackage{blindtext}

\usepackage{jmlr2e}

\usepackage{lastpage}
\usepackage{booktabs}
\usepackage{array}
\usepackage{subcaption}
\usepackage{amsmath}
\usepackage{hyperref}

\firstpageno{1}

\begin{document}

\title{Improving satellite imagery segmentation using multiple Sentinel-2 revisits}

\author{\name Kartik Jindgar \email kartik.jindgar@nyu.edu \\
       \addr Center for Data Science\\
       New York University\\
       New York, NY 10011, USA
       \AND
       \name Grace W. Lindsay \email grace.lindsay@nyu.edu \\
       \addr Center for Data Science\\
       New York University\\
       New York, NY 10011, USA}
       
\editor{N/A}

\maketitle

\begin{abstract}
In recent years, analysis of remote sensing data has benefited immensely from borrowing techniques from the broader field of computer vision, such as the use of shared models pre-trained on large and diverse datasets. However, satellite imagery has unique features that are not accounted for in traditional computer vision, such as the existence of multiple revisits of the same location.
Here, we explore the best way to use revisits in the framework of fine-tuning pre-trained remote sensing models. We focus on an applied research question of relevance to climate change mitigation---power substation segmentation---that is representative of applied uses of pre-trained models more generally. Through extensive tests of different multi-temporal input schemes across diverse model architectures, we find that fusing representations from multiple revisits in the model latent space is superior to other methods of using revisits, including as a form of data augmentation. We also find that a SWIN Transformer-based architecture performs better than U-Nets and ViT-based models. We verify the generality of our results on a separate building density estimation task.  

\end{abstract}


\section{Introduction}

With the establishment of several government-run satellite projects, numerous global and regularly updated public sources of remote sensing data are becoming readily available. This abundance, coupled with advancements in  computer vision models, has fueled a growing interest in applying advanced machine learning and computer vision techniques to remote sensing and earth observation data. Consequently, a wide selection of toolboxes \citep{Stewart_TorchGeo_Deep_Learning_2022}, datasets \citep{bigearth, satlas, phileo, eurosat, so2sat, torchgeo_dataset} and pretrained-models \citep{satlas, Prithvi, Stewart_TorchGeo_Deep_Learning_2022} have been released to support this field.

These models have applications across many fields but particularly in addressing climate change. For instance, accurately mapping roads and buildings \citep{road_detection1, road_detection2, building_segmentation1, building_segmentation2} can help in urban planning, population mapping and disaster response. Monitoring changes in land use \citep{land1, land2} can help track urban expansion and deforestation. Detecting vessel positions and types \citep{shiping1, shipping2} in the sea can aide identification of illegal fishing, exploration and mining. Crop yield prediction \citep{yield_prediction} can help address crop production challenges caused by climate variability.  Finally, tracking natural disasters \citep{flooddetection,wildfiredetection,naturaldisaster} can be vital for improving emergency response and minimizing loss to human life and property. Despite these recent advances, many relevant climate change problems are still not fully benefiting from remote sensing data.

Remote sensing tasks have largely relied on general computer vision models developed in other domains, yet significant differences exist between satellite imagery and other computer vision datasets \cite{rolf2024mission}. One of the most unique features of remote sensing data is the inclusion of multiple revisits -- i.e. images taken at the same place at different points in time.  The temporal resolution of a satellite determines the frequency of these revisits, which can vary widely. For example, each single Sentinel-2 satellite revisits a location in 10 days, while the combined constellation reduces this interval to 5 days. Even though images from all the revisits might not be useful due to excessive cloud cover, this temporal dimension introduces opportunities. Previous works have used revisits in various ways, including as data augmentation \citep{data_aug} and to synthesize features across time \citep{mt-attention, mt-sum-final-layer}. 

In this work we seek to identify the most effective use of revisits in the context of modern pre-trained remote sensing models. Taking an application-driven approach \citep{rolnick2024application}, we explore this question using a building segmentation problem of relevance to the energy sector: segmentation of power substations. Our power substation dataset was collected by TransitionZero as part of its efforts to create open and accessible data on the energy transition, particularly in data-sparse regions. Through comparing multiple approaches across several common architectures, we find that combining information from multiple revisits in the latent space significantly increases performance. To verify the robustness and the generality of our findings, we replicate these results in a separate building density estimation task. Our work provides valuable insights for the remote sensing community, providing a simple yet highly effective approach to enhance performance through the use of revisits.

\section{Related Work}
The temporal dimension of satellite imagery provides a unique opportunity to learn better representations. Previous works have leveraged revisits in many unique and interesting ways. One straightforward use is as a form of data augmentation; adding multiple revisits to the data set has been shown to increase performance on building and road extraction tasks \citep{data_aug}. \cite{seco},  \cite{mt-contrastive-learning}, and \cite{mt-contrastive-learning2} use revisits in a contrastive self-supervised learning framework to derive robust representations that improve accuracy on downstream tasks. These methods learn by enforcing representations of the same location, across time, to be closer than those of different locations. \cite{mt-concatenate} stack time series images per crop-growing season to construct a single image before passing it through a Trans U-Net, thus combining revisits in the input space. Other methods have synthesized latent features across revisits. \cite{mt-attention} apply self-attention masks to aggregate temporal features in a panoptic segmentation task. \cite{satlas} fuses revisits in the latent feature space using a max operation for a variety of downstream tasks. \cite{mt-sum-final-layer} sums individual predictions from each timepoint to make the final prediction for a functional use task where changes over time are expected to reveal building use. Finally, it is also possible to use multiple timepoints from low resolution imagery to generate a high resolution input. However, \cite{super-resolution} provide evidence that for downstream tasks such as segmentation, super resolved images do not outperform original low resolution images. 

Here, we aim to determine how revisits can best be used along with pre-trained models that may or may not have been designed to utilize revisits. To do so, we develop new architectural alterations for U-Net and ViT-based models and replicate an existing SWIN-based model \cite{satlas}. Importantly, the architectures we test can work in the context of any number of (including only one) revisits and can apply to tasks without an obvious temporal component; this allows flexibility and does not overly constrain the type of data that can be used. The need to use revisits with pre-trained models is a realistic scenario in applied research settings where users want to get the best performance on their specific task by starting with general purpose pre-trained models. Therefore identifying simple yet highly effective techniques for combining temporal features is of value for remote sensing applications across a variety of domains.  


\begin{figure*}[t!]
    \centering
    \begin{subfigure}[t]{1\linewidth}
        \centering
        \includegraphics[height=1.2in]{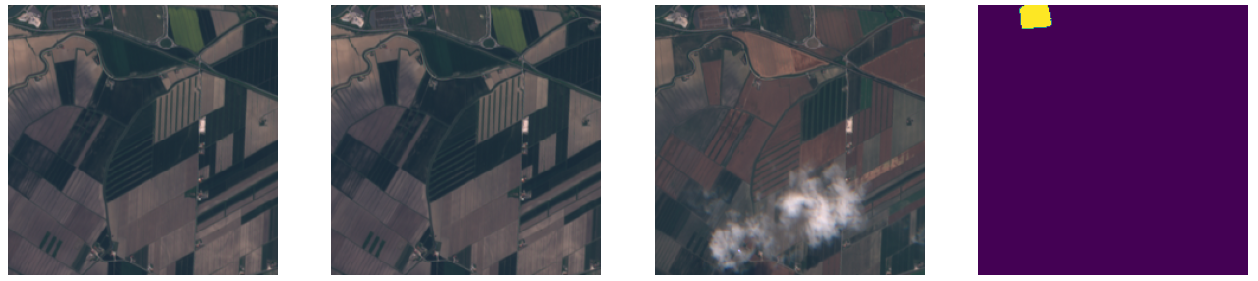}
        \caption{Power-Substation Dataset}
    \end{subfigure}
    \\
    \begin{subfigure}[t]{1\linewidth}
        \centering
        \includegraphics[height=1.2in]{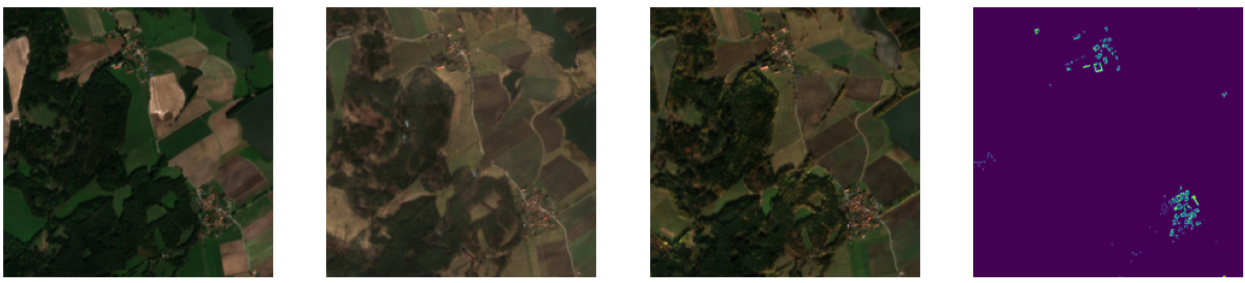}
        \caption{PhilEO Building Dataset}
    \end{subfigure}
    \caption{Images from multiple revisits along with mask}
    \label{fig:sample_input}
\end{figure*}

\section{Methodology}
\subsection{Dataset} \label{dataset}
We ultimately employ two unique geospatial datasets in our experiments to compare the performance of different segmentation models. The first dataset is curated by TransitionZero and sourced from publicly available data repositories, including OpenSreetMap \citep{openstreet} and Copernicus Sentinel data. The dataset consists of Sentinel-2 images from 27k+ locations; the task is to segment power-substations, which appear in the majority of locations in the dataset. Most locations have 4-5 images taken at different timepoints (i.e., revisits) and each image is of dimension 228x228 pixels. Each image has 13 spectral bands and each band has been linearly interpolated to a spatial resolution of 10m. Lastly, there is one ground truth mask for each location. 

The PhilEO Downstream Dataset \citep{phileo} is also a global Sentinel-2 dataset with 11 bands at 10m resolution. The bands include a Scene Classification Layer and 10 Sentinel-2 spectral bands (note: the Scene Classification Layer is not used here). Each location has been revisited at least 3 times (to achieve this we filtered out locations with less than 3 revisits). The dataset originally comprises three tasks: building density estimation, road segmentation, and land cover classification. We test the models on building density estimation because of its similarity to power substation segmentation. 

Both datasets were pre-processed to remove images with cloud cover.  The substation dataset was divided into training and testing sets with an 80-20 split, resulting in images from 22k and 5k locations for training and testing respectively. For the PhilEO dataset, the training set consisted of images from 50k locations, with images from about 6k locations heldout for testing. We evaluated different normalization techniques including scaling by a constant(e.g. 4000), computing z-scores and normalization using 1st and 99th percentiles. We found that different strategies worked best for different pre-trained models (Refer to Appendix \ref{app:normalization}). Furthermore, we applied geometric augmentations to training images, such as random affine transformations, random flips and random rotations.

\subsection{Multi-Temporal Input}\label{multi-temp-input}
As mentioned above, a unique feature of satellite imagery is its multi-temporal aspect. Satellites generate multi-temporal images because they continuously orbit the Earth, re-visiting the same location repeatedly. This results in a unique problem and opportunity for computer vision models using satellite imagery. We identify and evaluate five different general strategies to handle this multi-temporal input in the context of fine-tuning pre-trained models. 
\begin{itemize}
    \item \textbf{Single-Image Input}: Use just one image from all the revisits and discard the rest. This would result in a more orthodox dataset that is similar to other computer vision domains.
    \item \textbf{Augmented Single-Image Input}: Use each revisit separately to augment the dataset. This means multiple different images are associated with the same segmentation map, but treated as separate images during training. As shown by \cite{data_aug}, this increases the size of the dataset while preventing undesired synthetic artifacts that may arise from photometric data augmentation techniques; in our case the dataset becomes 3-4 times the size of the original.
    \item \textbf{Averaged Single-Image Input}: Compute an average (such as the median) image of all the revisits and use that for training. As with the first method, this results in a single image per location, but unlike the first method it still utilizes data from all the revisits.
    \item \textbf{Latent Fusing of Multi-Image Input}: Generate individual embeddings for each image and fuse them in the model's latent space before passing to the decoder. This approach effectively utilizes information from each revisit at each training step. \textit{Demonstrating the superiority of this method is our main contribution}. Through preliminary investigations, we found that the temporal max operation performs best. Details of how this fusion is done for different model architectures are given in the following section. 
    \item \textbf{Output Fusing of Multi-Image Input}: Generate individual pre-threshold segmentation maps for each revisit, then threshold the median of these images. This method also utilizes all revisits for each training step, but does not require alterations to the internal model layers and is therefore more agnostic to model architecture.  
\end{itemize}

\begin{figure}[]
    \centering
    \includegraphics[width=1\linewidth]{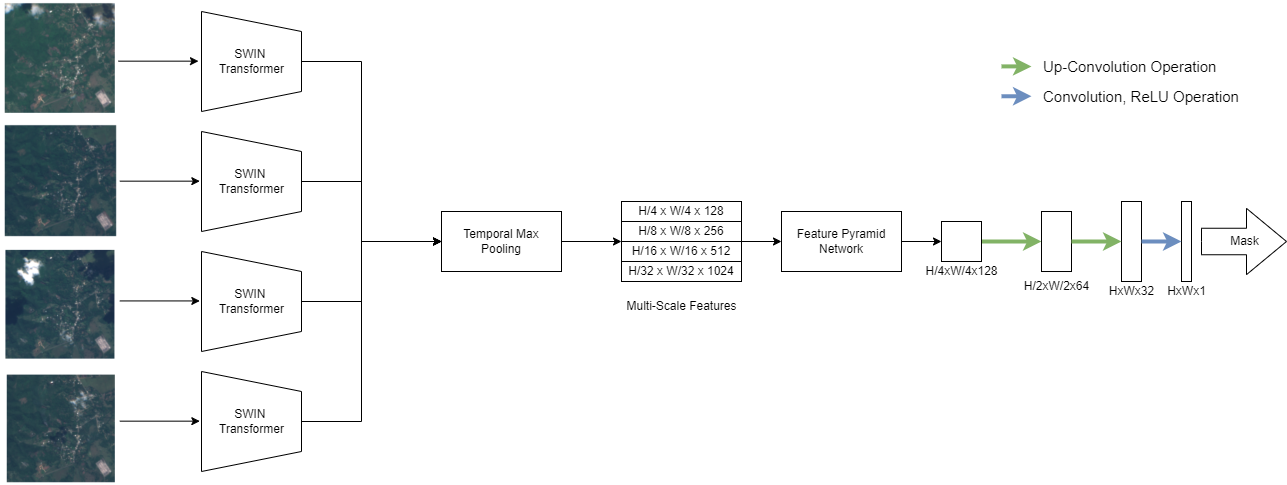}
    \caption{Segmentation model with SWIN transformer backbone and multi-temporal input}
    \label{fig:swin_arch}
\end{figure}

\begin{figure}[]
    \centering
    \includegraphics[width=0.8\linewidth]{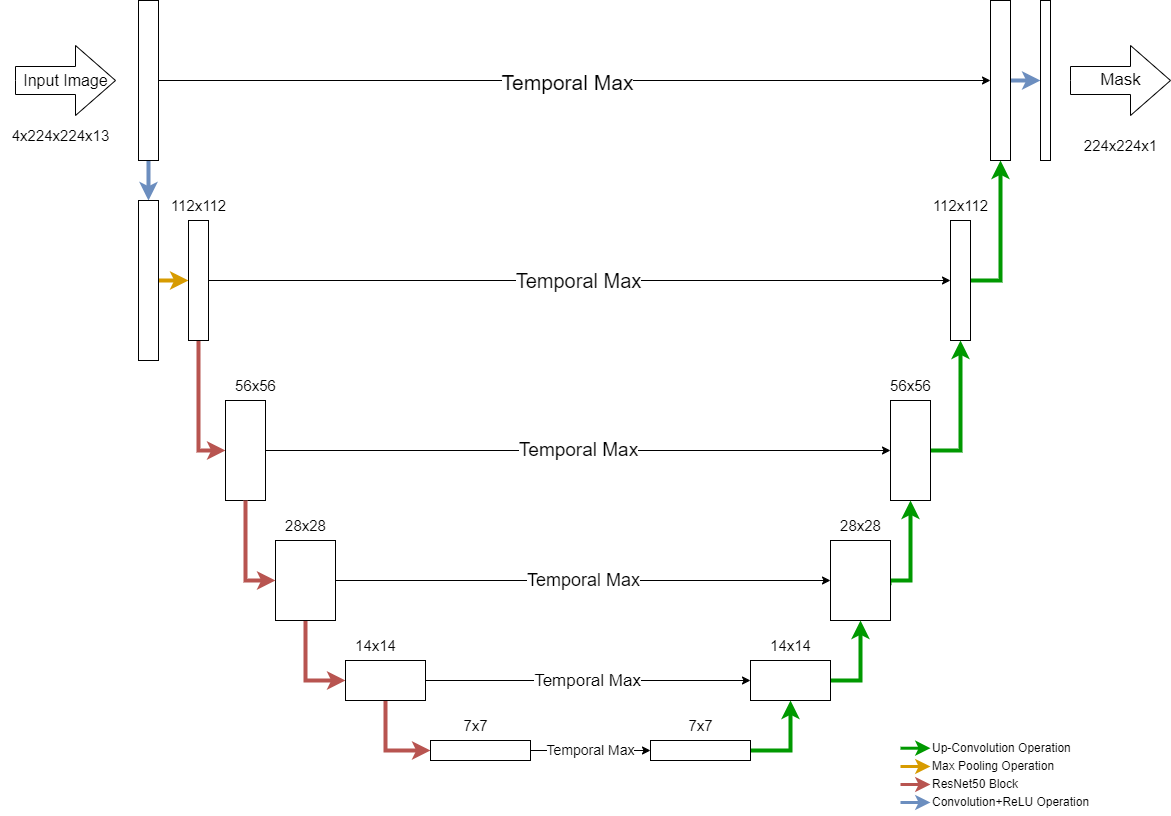}
    \caption{U-Net with a ResNet50 backbone and multi-temporal input}
    \label{fig:U-Net_arch}
\end{figure}

\subsection{Model Architectures}
We experimented with three different kinds of segmentation models\footnote{The code is available \href{https://github.com/Lindsay-Lab/substation-seg/}{here}}. These included a U-Net \citep{unet} with a ResNet50 backbone \citep{resnet}, a model with a SWIN Transformer backbone \citep{swin} and a model with a ViT backbone \citep{vit}. In all the models, the encoder was pre-trained while the decoder was randomly initialized. Furthermore, the encoder was not frozen and was allowed to fine-tune during training. All architectures were adapted (as necessary) to handle multi-temporal inputs, as described below.

The ResNet50 backbone in the U-Net, provided by Torchgeo \citep{Stewart_TorchGeo_Deep_Learning_2022}, was pre-trained on the SSL4EO-S12 satellite imagery dataset \citep{torchgeo_dataset} (which contains all 13 Sentinel-2 bands) using the MoCo self-supervised learning method \citep{moco}. To test on our datasets, input images (containing all 13 bands) were rescaled to a size of 224x224 pixels. To adapt the U-Net for multi-temporal input, individual timepoints were passed through the ResNet encoder and the features at each scale were fused using a temporal max operation before being passed to the decoder (Figure \ref{fig:U-Net_arch}). 

For the SWIN-based models, the SWIN transformer backbone was pretrained on the SatlasPretrain dataset \citep{satlas} using supervised learning. SatlasPretrain is a large scale remote sensing dataset that combines 9-band Sentinel-2 images (see Appendix Table \ref{tab: ms input} for bands) and NAIP images annotated with 302M distinct labels. For fine-tuning, input images of size 228x228 were passed through the encoder which generated features at 4 different scales - H/4xW/4, H/8xW/8, H/16xW/16, H/32xW/32. Similar to the U-Net and as done by \cite{satlas}, for multi-temporal input, features at each scale were fused together using a temporal-max operation (Figure \ref{fig:swin_arch}). These features were then processed by an FPN network followed by upsampling blocks. Each upsampling block consisted of ConvTranspose, Conv2d and ReLU layers.

For the ViT-based model, we explored several pre-trained backbones. First, we used three differently-sized (small, base, and large) ViT models from the PyTorch Image Models library \citep{timm}. These took RGB input only and were pre-trained on ImageNet \citep{imagenet} in a supervised fashion. Additionally, a small multi-spectral ViT backbone from TorchGeo was used, which was pretrained on SSL4EO-S12 using the DINO self-supervised learning method \citep{dino}. Our preliminary experiments revealed that the base ViT encoder pretrained on ImageNet outperformed all the others (see Appendix \ref{vit-encoder-size}). This performance difference was possibly due to the larger size of the encoder pretrained on ImageNet compared to the multi-spectral network (TorchGeo does not currently offer a larger pre-trained multispectral ViT). Consequently, all subsequent experiments involving the ViT model were conducted using the base RGB-only ViT encoder pretrained on ImageNet (note that for the U-Net and SWIN models, restricting to RGB input only negatively impacted performance as seen in Appendix Table \ref{ms-input})). For multi-image inputs to the ViT model, latent embeddings of each individual timepoint are generated by the ViT transformer backbone. These are fused using a temporal max operation before being passed to the decoder. Out of the different decoder architectures tested, the simplest one worked the best. It is composed of sequential upsampling blocks that transform the feature embedding from the encoder into an output mask of shape 224x224(Appendix \ref{app:vit_decoder}). Each block was composed of ConvTranspose, Conv2d, BatchNorm, Dropout and ReLU layers. As was the case with the U-Net, the input images were rescaled to 224x224.


\begin{table}
\centering
\begin{tabular}{ccccc} 
\\\toprule
Model & \multicolumn{1}{c}{\begin{tabular}[c]{@{}c@{}}Pretrained\\ Encoder\end{tabular}} & Channels & Timepoints &   IoU \\ \midrule
ViT   &   Yes (Imagenet)   &    3     &    4       &  0.41	$\pm$ 0.009 \\
U-Net  &  Yes  (Torchgeo)   &    13    &    4       &  0.48 $\pm$ 0.004\\
SWIN  &   Yes (SATLAS)     &    9     &    4       &  \textbf{0.58  $\pm$ 0.003}\\
\bottomrule
\end{tabular}
\caption{Comparison of different pretrained models on power-substation segmentation task based on Intersection over Union (IoU). Temporal max operation is used to fuse features of each timepoint and geometric augmentations are applied on input images. }
\label{overall-iou}
\end{table}

\begin{table}
\centering
\begin{tabular}{ccccc}
\\\toprule
Model &  Pre-trained Encoder &       Channels       &  Timepoints     & IoU \\ \midrule
ViT   &      No              &       3              &     4   & 0.16 $\pm$ 0.009 \\
U-Net  &      No              &       13             &     4  &0.45 $\pm$ 0.007 \\
SWIN  &      No              &       9              &      4 & \textbf{0.48 $\pm$ 0.011} \\
\bottomrule
\end{tabular}
\caption{Comparison of different models, trained from scratch, on power-substation segmentation task. Temporal max operation is used to latently fuse features of each timepoint and geometric augmentations are applied on input images. }
\label{pre-trained}
\end{table}

\section{Results}
We conducted experiments to evaluate the performance of three architectures -- SWIN, ViT and U-Net -- along with their respective pre-trained weights on the substation segmentation task outlined in Section \ref{dataset}. Of particular interest was the question of which of the possible methods for utilizing revisits (Section \ref{multi-temp-input}) leads to highest performance. We also performed additional experiments to show the effectiveness of pre-training. We then show how our results replicate on the PhilEO building density estimation task (see Section \ref{dataset}). For all the experiments, we report mean and standard error over 5 runs with random seeds. 

\subsection{SWIN outperforms ViT and U-Net}\label{subsec:swin_vs_vit_vs_unet}
Table \ref{overall-iou} presents a comparison of the three pretrained models (all using multiple revisits fused in the latent space as described above) on the substation segmentation task. Our analysis reveals that the SWIN model comprehensively outperforms both the ViT and U-Net models. Notably, the U-Net outperforms the ViT model, likely due to the absence of skip connections in the ViT model. Without skip connections, the amount of information used by the decoder is limited and restricts the quality of image reconstruction. Both the U-Net and SWIN models use features at different scales, which overcomes this hurdle. Our results are broadly consistent with previous work that has explored similar base architectures \citep{lacoste2024geo}.

We also evaluated the fine-tuned SWIN-based model on a separate dataset of negative images (i.e. images without substations) to assess if it incorrectly predicted a substation in these images. Our experiments revealed that the SWIN model misclassified pixels in less than 5\% of the 19k negative images. In most of these incorrectly-labeled images, the model labeled fewer than 100 pixels as substation out of 228x228 total pixels. 

In this work, we are motivated to use pre-trained models under the assumption that training on a larger database will create better representations for our specific task. To test this assumption directly, we also trained our models from scratch on our dataset. This comparison also isolates the impact of architecture on performance, as it removes differences in pre-training data and methods that exist across our different models. Furthermore, a concern when evaluating pre-trained remote sensing models is the possibility of overlap between training images and testing images; this is possible specifically for our U-net and SWIN models as both are pre-trained on Sentinel-2 data (and the SWIN model's pre-training task included substations as one of its many types of labels). As shown in Table \ref{pre-trained}, even when trained from scratch, the SWIN model outperforms both the U-Net and ViT models, underscoring its superiority over the other architectures. In terms of benefits of pre-training, the ViT model takes the biggest performance hit when training from scratch, while the U-net is least impacted.  

\begin{table}
\centering
\begin{tabular}{ccccc}
\\\toprule
Model &  Temporal Dimension &  Temporal Aggregation & IoU \\ \midrule
ViT  &     1       &       Single Image     &        0.36 $\pm$ 0.010 \\
ViT  &     1       &       Augmented Dataset     &     0.37 $\pm$ 0.010\\
ViT  &     1       &       Median Image         &  0.38 $\pm$ 0.006 \\
ViT  &     4       &       Output Fusion         &  0.39 $\pm$ 0.007 \\
ViT  &     4       &       Latent Temporal Max         &\textbf{0.41	$\pm$ 0.009}\\
\hline
U-Net   &    1       &       Single Image     &        0.41 $\pm$ 0.013 \\
U-Net  &     1       &       Augmented Dataset     &     0.41 $\pm$ 0.016\\
U-Net  &     1       &       Median Image         &  0.44 $\pm$ 0.003 \\
U-Net  &     4       &       Output Fusion         &  0.47 $\pm$ 0.008 \\
U-Net  &     4       &       Latent Temporal Max         &\textbf{0.48	$\pm$ 0.004}\\
\hline
SWIN  &     1       &       Single Image     &        0.55 $\pm$ 0.001 \\
SWIN  &     1       &       Augmented Dataset     &     0.55 $\pm$0.003\\
SWIN  &     1       &       Median Image         &  0.56 $\pm$0.002 \\
SWIN  &     4       &       Output Fusion         &  0.56 $\pm$0.002 \\
SWIN  &     4       &       Latent Temporal Max         &\textbf{0.58	$\pm$ 0.003}\\
\bottomrule
\end{tabular}
\caption{Comparison of different strategies for utilizing multi-temporal input on power-substation segmentation task. All the models used a pretrained encoder and respective multi-spectral inputs (Appendix \ref{app:ms_input}).}
\label{time-iou}
\end{table}

\subsection{Incorporating multiple revisits improves performance}\label{subsec:mt_data}
Our main question of interest is how to best utilize revisits as a unique feature of satellite imagery. We conducted experiments to evaluate and compare the different techniques of handling multi-temporal input as defined in Section \ref{multi-temp-input}. Table \ref{time-iou} shows performance for all five revisit techniques for each of the three pre-trained models.  Combining the latent intermediate features using a temporal max operation results in a 14\%, 17\% and 5\% lift in performance (over the single-image baseline) for the ViT model, U-Net and SWIN model respectively. Additionally, we find that computing a median image from all the revisits also improves the performance by a few points as compared to using a single image. Interestingly, using each single revisit to augment the dataset doesn't lead to any noticeable improvement in performance over using just any one single image out of all the revisits; this is in contrast to what was found by \cite{data_aug}.


\begin{table}
\centering
\begin{tabular}{ccccc}
\\\toprule
Model &  Temporal Dimension &  Temporal Aggregation & MSE \\ \midrule
ViT  &     1       &       Single Image     &        0.0044 $\pm$ 0.00010 \\
ViT  &     1       &       Augmented Dataset     &     0.0044 $\pm$ 0.00003\\
ViT  &     1       &       Median Image         &  0.0043 $\pm$ 0.00005 \\
ViT  &     4       &       Output Fusion         &  0.0045 $\pm$ 0.00002 \\
ViT  &     4       &       Latent Temporal Max         &\textbf{0.0042	$\pm$ 0.00004}\\
\hline
U-Net   &    1       &       Single Image     &        0.0033 $\pm$ 0.00008 \\
U-Net  &     1       &       Augmented Dataset     &     0.0033 $\pm$ 0.00003\\
U-Net  &     1       &       Median Image         &  0.0032 $\pm$ 0.00008\\
U-Net  &     4       &       Output Fusion         & 0.0040	$\pm$ 0.00012\\
U-Net  &     4       &       Latent Temporal Max         & \textbf{0.0029 $\pm$ 0.00004}\\
\hline
SWIN  &     1       &       Single Image     &        0.0023 $\pm$ 0.00001 \\
SWIN  &     1       &       Augmented Dataset     &     0.0024 $\pm$ 0.00002\\
SWIN  &     1       &       Median Image         &  0.0022 $\pm$0.00003 \\
SWIN  &     4       &       Output Fusion         &  0.0024	 $\pm$0.00003 \\
SWIN  &     4       &       Latent Temporal Max         &\textbf{0.0019	$\pm$ 0.00001}\\
\bottomrule
\end{tabular}
\caption{Comparison of different strategies for utilizing multi-temporal input on building density estimation task based on Mean Square Error(MSE). All the models used a pretrained encoder and respective multi-spectral inputs (Appendix \ref{app:ms_input}).}
\label{time-mse}
\end{table}

\subsection{PhilEO-Downstream Dataset}
To test the generality of our claims, we ran the experiments outlined in Sections \ref{subsec:swin_vs_vit_vs_unet} and \ref{subsec:mt_data} on the PhilEO Downstream dataset and show results in table \ref{time-mse}. Consistent with our earlier findings, the SWIN model demonstrated superior performance compared to the U-Net and ViT models on the building density estimation task. Once again, the ViT model ranked worst amongst the three models. Also aligned with our earlier results, the latent temporal fusion strategy proved to be the most effective strategy of incorporating multi-temporal images, resulting in a 5\%, 12\% and 17\% improvement in the ViT, U-Net, and SWIN model's performance respectively compared to using a single image as input. Similarly, using median images led to some improvement over single image inputs. As with our substation results, the output fusion method performed inconsistently, occasionally performing even worse than the single image baseline (though note we did experience optimization issues with this approach, especially when using the U-net architecture). Finally, our best models outperform models from \cite{phileo}, further validating the effectiveness of the temporal fusion strategy (though note the comparison is not exact because we are using a restricted dataset of only locations that contained at least 3 revisits). In total, we find that combining temporal data in the latent space is a universally effective and robust way to utilize revisits in satellite imagery.

\begin{figure}[]
    \centering
    \includegraphics[width=0.8\linewidth]{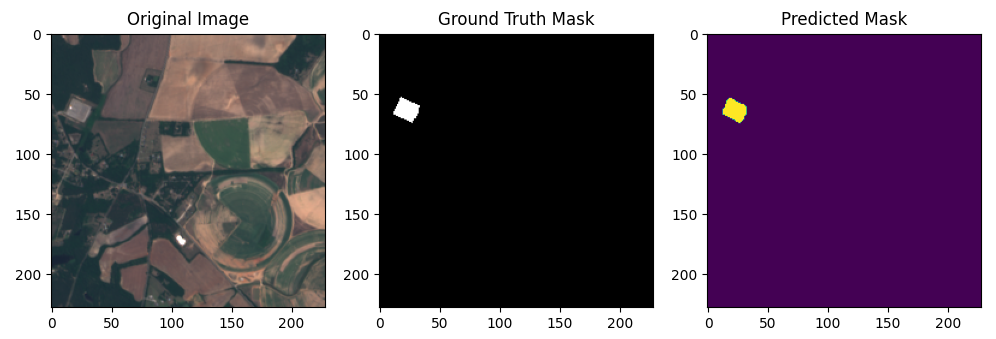}
    \\[\smallskipamount]
    \includegraphics[width=0.8\linewidth]{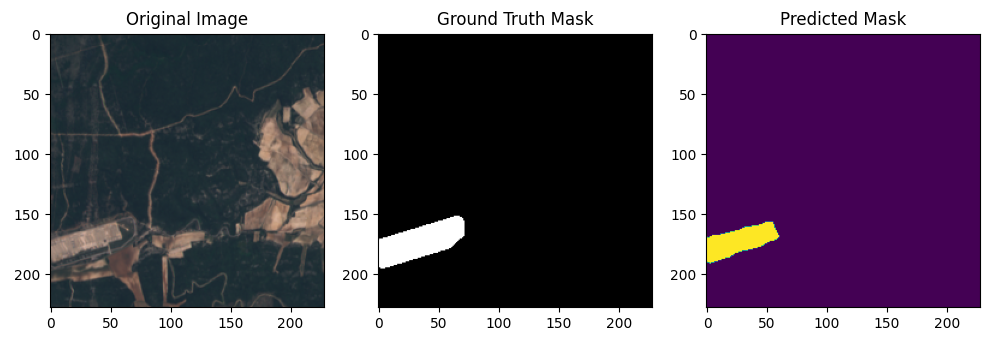}
    \caption{Predictions by the SWIN model on the power-substation dataset }
    \label{fig:sample output}
\end{figure}

\section{Conclusion}
In this study we set out to find the most effective way to utilize revisits in the framework of pre-trained remote sensing models. Through our extensive experiments we found that fusing representations from multiple revisits in the latent space significantly improves performance. This general principle held across architectures and datasets.
Additionally, we observed that models with a SWIN transformer backbone comprehensively beat other architectures.

Importantly, we took an application-driven approach to this problem by focusing on a dataset and task that is of interest to the energy sector, particularly in planning the energy transition \citep{rolnick2024application}. The problems faced in this application are common for many use-cases for remote sensing and geospatial data: a specific task that is not necessarily represented in many existing benchmarks, a small- to medium-sized dataset, and the desire to minimize compute costs. These constraints create a setting where fine-tuning pre-trained models is a natural solution. By identifying the best way to utilize revisits in the context of fine-tuning specific pre-trained remote sensing models, we believe we are providing valuable insights for many applied researchers.

We also believe our work should drive future studies exploring the most effective ways of fusing representations in the latent space, as we only explored a small subset of possible ways here. Another interesting question is that of how many revisits are necessary, and does performance plateau after a certain number of revisits (when using revisits as data augmentation, \cite{data_aug} saw performance increase as a function of revisits up to 3). Finally, we expect our findings to apply to tasks involving static features such as buildings or roads. However it is possible that revisits taken within a short period of time (e.g., the multiple images Sentinel-2 takes in the course of a month) may help even in the context of more temporally varying targets.  

\acks{We would like to thank Joe O'Connor and Lucas Kruitwagen from TransitionZero for providing the substation dataset and giving input on this project. We would also like to thank Simran Makariye and Pooja Aryamane for their contributions to the work. GL and KJ declare no competing interests.}


\newpage

\appendix

\begin{table}
\centering
\begin{tabular}{ccc}
\\\toprule
Model &  Normalization Technique & IoU \\ \midrule
SWIN   &      Normalization              &        0.572 $\pm$ 0.0033 \\
SWIN  &      Scaling with a constant     &       \textbf{0.582 $\pm$ 0.0026} \\
\bottomrule
\end{tabular}
\caption{Comparison of different data pre-processing techniques for the SWIN model on power-substation segmentation task. Temporal max operation is used to latently fuse features of each timepoint. }
\label{tab:swin-preprocessing}
\end{table}

\begin{table}
\centering
\begin{tabular}{ccc}
\\\toprule
Model &  Normalization Technique & IoU \\ \midrule
ViT   &      Scaling with a constant  &        0.335 $\pm$ 0.0001 \\
ViT  &      Standardization     &       \textbf{0.382	 $\pm$ 0.0001} \\
\bottomrule
\end{tabular}
\caption{Comparison of different data pre-processing techniques for the ViT based model on power-substation segmentation task. Single image input is used for this experiment. }
\label{tab:vit-preprocessing}
\end{table}

\section{Data Pre-processing Techniques} \label{app:normalization}
We experimented with three different data pre-processing techniques for scaling pixel values of input images: standardization, normalization and scaling down by a constant. 
\begin{itemize}
    \item Standardization:
    
            \begin{equation}        
                \text{z-score} = \frac{\text{input} - \text{mean}}{std}
            \end{equation}

    \item Normalization: 

            \begin{equation}
                \text{normalized\ value} = \frac{input - min} {max - min}
            \end{equation}

    \item Scaling down by a constant:
    
            \begin{equation}
                new\ value = \text{clip}\left(\frac{input}{constant}, 0, 1\right)
            \end{equation}
    
\end{itemize}

For standardization, we used a large random sample of the dataset to compute the mean and standard deviation. Since normalization is sensitive to outliers, we used the 1st and 99th percentiles in place of minimum and maximum values. Scaling down by a constant, which in this case is similar to normalization since our input image data didn't have negative values, worked best with a constant value of 4000. Different strategies worked well for different pretrained models. For the SWIN model and the U-Net, scaling with a constant yielded the best results(see SWIN results in table \ref{tab:swin-preprocessing}), while for the ViT model, standardization performed the best(see table \ref{tab:vit-preprocessing}). This was possibly because the ViT encoder was pretrained on ImageNet where z-score standardization is applied on input data.

\section{Multi-spectral Input}\label{app:ms_input}

\begin{table}
\centering
\begin{tabular}{cccccccccccccc}
\\\toprule
Sentinel 2 Bands & B1 &  B2 & B3 & B4 & B5 & B6 & B7 & B8 & B8a & B9 & B10 & B11 & B12 \\ \midrule
U-Net & \checkmark & \checkmark & \checkmark & \checkmark & \checkmark & \checkmark & \checkmark & \checkmark & \checkmark & \checkmark & \checkmark & \checkmark & \checkmark \\
SWIN & & \checkmark & \checkmark & \checkmark & \checkmark & \checkmark & \checkmark & \checkmark & & & & \checkmark & \checkmark \\
ViT &  & \checkmark & \checkmark & \checkmark &  & &  &  &  &  &  &  &  \\
\hline
\end{tabular}
\caption{Composition of multi-spectral input for different models}
\label{tab: ms input}
\end{table}

Throughout this work we utilized different sets of  Sentinel-2 spectral bands for fine-tuning different pre-trained models. Specifically, the multi-spectral input to the U-Net, SWIN and ViT models consisted of 13, 9 and 3 bands, respectively. As discussed earlier, we used the ViT encoder that was pre-trained on ImageNet. As a result, the ViT based model was fine-tuned with only RGB inputs. In contrast, the U-Net and SWIN models, were trained with additional spectral bands alongside the RGB channels (Table \ref{tab: ms input}). 

The PhilEO dataset lacked three of the 13 spectral channels required for fine-tuning the U-Net -- B1, B9 and B10 channels. To address this, we substituted the missing channels with existing bands that were highly correlated. Specifically, B1 was replaced by B2, B9 by B8A, and B10 by B11.

\section{Incorporating more channels boosts performance}
\label{rgb_vs_ms}

\begin{table}
\centering
\begin{tabular}{ccccc}
\\\toprule
Model &     Channels       &        IoU \\ \midrule
U-Net  &       3            &       0.33 $\pm$ 0.0025 \\
U-Net  &       13           &       0.44 $\pm$ 0.0029\\
SWIN  &       3            &       0.56 $\pm$ 0.0031 \\
SWIN  &       9            & \textbf{0.58	$\pm$ 0.0026}\\

\hline
\end{tabular}
\caption{Comparison between multi-spectral and RGB inputs for power-substation segmentation task. Both the U-Net and SWIN model consisted of pre-trained encoders and used a temporal-max operation to utilize the multi-temporal inputs.}
\label{ms-input}
\end{table}

We also carried out an experiment to understand the impact of incorporating additional channels in the U-Net and SWIN models for the power-substation segmentation task. Table \ref{ms-input} shows that multi-spectral configurations of both these models perform much better than RGB ones.  

\begin{table}
\centering
\begin{tabular}{ccccc}
\\\toprule
Encoder Size  &   Pre-trained       & Channels &      IoU \\ \midrule
Small  &       No            &     3 & 0.139  \\
Small  &       Yes(ImageNet)  &     3    &       0.291 \\
Small  &       Yes(TorchGeo)  &     13    &       0.294 \\
\hline
Base  &       No            &  3  &   0.150\\
Base  &       Yes(ImageNet)  & 3         &       \textbf{0.335} \\
\hline
Large  &       No            &  3 &     0.150 \\
Large  &       Yes(ImageNet)  & 3         &       0.300 \\
\hline
\end{tabular}
\caption{Comparing different ViT encoders on the power-substation segmentation task. Single image input was used and images were scaled with a constant(Appendix \ref{app:normalization})}
\label{tab:vit-encoder}
\end{table}

\section{Impact of the size of pre-trained ViT encoder on model performance}\label{vit-encoder-size}

Through our experiments we uncover an interesting relationship between the size of the ViT encoder and model performance. Contrary to expectations, using a larger encoder did not consistently lead to better results. Table \ref{tab:vit-encoder} shows that the Base ViT encoder performed the best amongst the three encoders, even better than the Large ViT encoder. Additionally, we also observed that using pre-trained encoders consistently resulted in significantly higher performance.   
\begin{figure}[]
    \centering
    \includegraphics[width=1\linewidth]{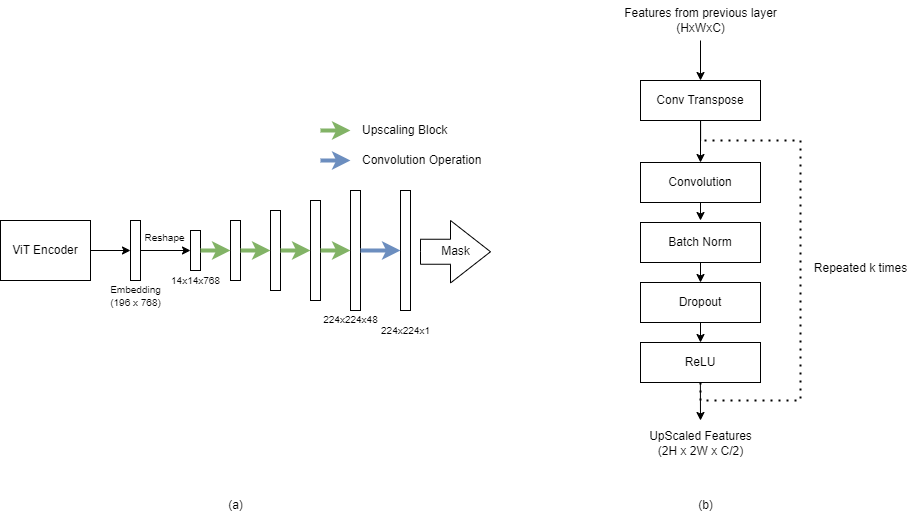}
    \caption{a) Architecture of decoder, comprising of multiple up-scaling blocks followed by a convolution layer, used in the ViT based model; b) Architecture of up-scaling block used in the decoder}
    \label{fig:vit-decoder}
\end{figure}

\section{ViT Decoder Architecture} \label{app:vit_decoder}
The ViT Encoder was applied in parallel to each timepoint and the individual embeddings were fused using a temporal max operation. The resulting embedding was passed to the decoder(Figure \ref{fig:vit-decoder}) which consisted of sequential up-scaling blocks that transformed the feature embedding into an output mask of shape 224x224. Each up-scaling block was composed of ConvTranspose, Conv2d, BatchNorm, Dropout and ReLU layers.

\vskip 0.5in
\bibliography{sample}

\end{document}